\documentclass[11pt,a4paper]{article}
\usepackage[hyperref]{ranlp2025}
\usepackage{times}
\usepackage{latexsym}

\usepackage{microtype}

\aclfinalcopy 
\usepackage{graphicx}

\usepackage{makecell}
\usepackage{csvsimple}
\usepackage{graphicx}%
\usepackage{multirow}%
\usepackage{amsmath,amssymb,amsfonts}%
\usepackage{amsthm}%
\usepackage{mathrsfs}%
\usepackage[title]{appendix}%
\usepackage{xcolor}%
\usepackage{textcomp}%
\usepackage{manyfoot}%
\usepackage{booktabs}%
\usepackage{algorithm}%
\usepackage{algorithmicx}%
\usepackage{algpseudocode}%
\usepackage{listings}%

\usepackage{amssymb,graphicx,stackengine,xcolor}

\usepackage{amsmath}
\newcommand*{\crosssymbol}{%
    \text{%
      \raisebox{1ex}{%
        \makebox[0pt][l]{%
          \rule[-.2pt]{.75ex}{.4pt}%
        }%
        \makebox[.75ex]{%
          \rule[-1ex]{.4pt}{1.5ex}%
        }%
      }%
    }%
}

\usepackage{ifthen}
\newcommand*{\GottBERT}[2]{\ifthenelse{\equal{#2}{last}}{$\mathrm{GottBERT}_{#1}^{\crosssymbol}$}{$\mathrm{GottBERT}_{#1}$}}
\newcommand*{\GottBERTf}[2]{\ifthenelse{\equal{#2}{last}}{$^{\mathrm{f}}\mathrm{GottBERT}_{#1}^{\mathrm{\crosssymbol}}$}{$^{\mathrm{f}}\mathrm{GottBERT}_{#1}$}}

\newcommand{\GottBERTt}{$\mathrm{GottBERT}$}

\title{GeistBERT: Breathing Life into German NLP}

\author{
 \textbf{Raphael Scheible-Schmitt\textsuperscript{1,2}} and
 \textbf{Johann Frei\textsuperscript{3}}
\\
\\
 \textsuperscript{1}School of Computation, Information and Technology, Technical University of Munich, Germany,\\
 \textsuperscript{2}IS\textsuperscript{2}E - Intelligent Systems, Science and Engineering, LIACC polo on Azores University, Ponta Delgada, Portugal,\\
 \textsuperscript{3}Chair of IT Infrastructure for Translational Medical Research,\\ Faculty of Applied Computer Science, University of Augsburg, Germany
\\
 \small{
   \textbf{Correspondence:} \href{mailto:raphael.scheible@tum.de}{raphael.scheible@tum.de}
 }
}

\date{}

\begin{document}
\maketitle
\thispagestyle{plain}
\begin{abstract}
Advances in transformer-based language models have highlighted the benefits of language-specific pre-training on high-quality corpora. In this context, German NLP stands to gain from updated architectures and modern datasets tailored to the linguistic characteristics of the German language.
GeistBERT seeks to improve German language processing by incrementally training on a diverse corpus and optimizing model performance across various NLP tasks.
We pre-trained GeistBERT using fairseq, following the RoBERTa base configuration with Whole Word Masking (WWM), and initialized from {\GottBERTt} weights. The model was trained on a 1.3 TB German corpus with dynamic masking and a fixed sequence length of 512 tokens. For evaluation, we fine-tuned the model on standard downstream tasks, including NER (CoNLL 2003, GermEval 2014), text classification (GermEval 2018 coarse/fine, 10kGNAD), and NLI (German XNLI), using $F_1$ score and accuracy as evaluation metrics.
GeistBERT achieved strong results across all tasks, leading among base models and setting a new state-of-the-art (SOTA) in GermEval 2018 fine text classification. It also outperformed several larger models, particularly in classification benchmarks. To support research in German NLP, we release GeistBERT under the MIT license.
\end{abstract}

\section{Introduction}\label{sec1}
The advancement of neural language modeling (LM) in natural language processing (NLP) has been driven by the development of contextual pre-trained word representations, particularly through transformer-based architectures. Models like Bidirectional Encoder Representations from Transformers (BERT) \citep{devlin_bert_2019} have significantly impacted the field by providing robust, generalized representations that can be fine-tuned for specific downstream tasks, enhancing performance across various NLP applications. While much of the early work focused on English and multilingual models, it has become clear that single-language models, particularly those trained on large, high-quality corpora, can outperform their multilingual counterparts when applied to their target language.

Building on this understanding, the German NLP community has seen the introduction of models like GottBERT~\citep{scheible-etal-2024-gottbert}, which leveraged the German portion of the OSCAR~\citep{ortiz_suarez_monolingual_2020} corpus to create a high-performance RoBERTa-based~\citep{liu_roberta_2019} model tailored specifically for the German language. However, as the field evolves, so too must the approaches to model training. Recent developments in pre-training methodologies, such as Whole Word Masking (WWM)~\citep{10.1109/TASLP.2021.3124365} and the availability of newer, more extensive corpora like OSCAR23~\citep{jansen2022perplexedqualityperplexitybasedmethod}, OPUS~\citep{tiedemann-2012-parallel}, and mC4~\citep{xue-etal-2021-mt5}, present opportunities to further refine and enhance German language models.

To fully leverage these developments for German NLP, we introduce GeistBERT, a German Enhanced Incremental Semantically Tuned BERT model. GeistBERT builds on the foundation laid by the best checkpoint of the filtered GottBERT model (i.e. \GottBERTf{base}{best}) through continued pre-training \citep{gururangan-etal-2020-dont}, extending it with modern German datasets including OSCAR23 and mC4 from CulturaX \citep{nguyen2023culturaxcleanedenormousmultilingual}, Wikipedia, and several OPUS corpora. Since CulturaX already applies both deduplication and filtering, it provides a strong backbone of high-quality German text, while the additional corpora enrich the model with broader linguistic and domain diversity. By introducing Whole Word Masking (WWM) and leveraging the scale and variety of these sources, GeistBERT seeks to establish a new benchmark for German language models, with improved performance across various NLP tasks.


Our contributions are as follows:
\begin{itemize}
    \item We incrementally trained GeistBERT on top of \GottBERTf{base}{best} using a combination of modern German corpora (OSCAR23, OPUS, mC4), OpenLegal and Wikipedia.
    \item We integrated WWM into the pre-training process to enhance the model's ability to capture semantic relationships within the German language.
    \item We provide GeistBERT as base model to the community, accessible under an open-source license for further usage.
\end{itemize}

GeistBERT represents a step forward in the development of German-specific transformer models, offering enhanced capabilities through modern training techniques and high-varying data.

\section{Related Work}
The rise of transformer-based models like BERT \citep{devlin_bert_2019} marked a major shift in NLP, enabling significant performance improvements. Originally introduced as an English model and later as a multilingual version (mBERT), BERT’s success led to monolingual adaptations tailored to specific languages. For German, models like GermanBERT\footnote{\url{https://www.deepset.ai/german-bert}} and dbmdz BERT\footnote{\url{https://huggingface.co/dbmdz/bert-base-german-uncased}} emerged, trained on datasets of 12GB–16GB, sourced from Wikipedia, news articles, and legal texts.

RoBERTa enhanced BERT by training on a larger 160GB corpus, optimizing the architecture, and removing next sentence prediction. This strategy was applied to other languages, resulting in models like CamemBERT \citep{martin_camembert_2020} for French and RobBERT~\citep{delobelle2020robbert} for Dutch, highlighting the benefits of large, diverse training corpora and the use of language-specific vocabularies.

In German NLP, GBERT and GELECTRA \citep{chan-etal-2020-germans} built on this progress by training on 145GB of the OSCAR corpus \citep{ortiz_suarez_monolingual_2020} and additional sources, surpassing earlier German BERT models. These advancements underscored the impact of larger, well-curated datasets on model performance.
GottBERT further extended this development as one of the first German RoBERTa models, trained on the German OSCAR corpus. Its results demonstrated the importance of data diversity but also noted that excessive data cleaning might reduce corpus variance and affect downstream performance.
GeistBERT refines this lineage by increasing data variance, optimizing pre-training strategies, and achieving strong performance without increasing model size, making it a robust and accessible model for German NLP.

\section{Methodology}

\subsection{Training Data and Pre-training}
Compared to {\GottBERTt}, GeistBERT was trained on a substantially larger corpus, totaling approximately 1.3TB of text data. Training data was shuffled to support uniform sampling and minimize order effects during pre-training. GeistBERT was pre-trained using the same byte-level BPE tokenizer as {\GottBERTt}, following the GPT-2 design with a vocabulary size of 52k. While the tokenizer architecture mirrors GPT-2, the vocabulary itself was trained from scratch on German text. fairseq \citep{ott_fairseq_2019} was employed to compute the binary format for pre-training. Unlike GottBERT’s TPU-based setup, which processed text as a continuous stream, GeistBERT’s GPU training respected natural sentence boundaries. This preserves linguistic structure during pre-training and avoids cutting sequences in the middle of sentences.

\begin{table*}[!htbp]
\caption{Overview of datasets used for training. The table lists the individual corpora, their sizes in gigabytes, their data sources, and whether they were deduplicated or filtered. The final corpus aggregates all listed datasets, resulting in approximately 1.3 TB of training data.}\label{data-table}
\centering\small
\begin{tabular}{lcccccc}
\hline 
\bf Corpus & \bf Documents & \bf Size (GB) & \bf Data Source & \bf Deduplicated & \bf Filtered \\ 
\hline
mC4 \& OSCAR23  & 6,064,736,930 & 1316.57 & CulturaX & Yes & Yes \\
\makecell[l]{ELRC-4244, ELRC-4240, ELRC-4258, \\ELRC-4217, ELRC-4189, ELRC-4171, \\ELRC-4149} & 14,919,003 & 2.34 & OPUS & Yes & No \\
ECB & 1,732,472 & 0.29 & OPUS & No & No \\
EUbookshop & 18,203,612 & 2.34 & OPUS & No & No \\
Europarl & 2,234,583 & 0.36 & OPUS & No & No \\
EuroPat & 19,387,517 & 3.52 & OPUS & No & No \\
OpenSubtitles & 41,612,280 & 1.35 & OPUS & No & No \\
TildeMODEL & 5,059,688 & 0.79 & OPUS & No & No \\
German Wikipedia & 4,767,776 & 7.23 & Wikipedia & No & No \\
OpenLegalData & 209,526 & 2.48 & OpenLegal & No & No \\
\textbf{Final corpus} & \textbf{6,172,863,387} & \textbf{1337.28} & &  &  \\
\hline
\end{tabular}

\end{table*}

Using fairseq, we pre-trained the GeistBERT model on a highly variant corpus consisting of 1.3TB plain text data on 8 NVIDIA A40 GPUs. The model was trained with the RoBERTa base architecture for 100k update steps using a batch size of 8k, initializing the weights with {\GottBERTf{base}{x}}. We largely adhered to RoBERTa's default training configuration \citep{liu_roberta_2019}, including dynamic masking, optimizer settings, and fixed sequence lengths (512 tokens). A 10k iteration warmup was applied, gradually increasing the learning rate to a peak of 0.0007, followed by a polynomial decay to zero.

\subsection{Downstream Tasks}
We fine-tuned pre-trained BERT models using Huggingface \citep{wolf_huggingfaces_2020} scripts, optimizing batch size and learning rate via grid search. NER and classification (CLS) tasks were trained for up to 30 epochs, while NLI tasks ran for up to 10 epochs using fairseq-adapted hyperparameters. Each task was executed 24 times with varied hyperparameters, selecting the best checkpoint based on the highest $F_{1}$ score (accuracy for NLI). Performance was evaluated analogously to \citet{scheible-etal-2024-gottbert} and compared with results from that study. 
The parameter search space used for the grid search is summarized in Table~\ref{tab:hyperparams}. All tasks were processed using two Nvidia RTX 3090 GPUs, leveraging Huggingface's Transformers library (v4.34.1).

\begin{table}[htb]
    \caption{Hyperparameters used in the grid search of the downstream tasks.}
    \label{tab:hyperparams}
    \centering
    \begin{tabular}{lc}
         \bfseries Parameter & \bfseries Values\\
         \hline
         Learning Rate & 5e-5, 2e-5, 1e-5, 7e-6, 5e-6, 1e-6 \\
         Batch Size & 16, 32, 48, 64 \\
         Epochs & 30 \\
         \hline
    \end{tabular}
\end{table}

\paragraph{NLI}

We evaluated NLI on the German XNLI dataset \citep{conneau-etal-2018-xnli}, an extension of MultiNLI \citep{williams-etal-2018-broad}, with 122k training, 2490 development, and 5010 test examples per language. Performance was measured by accuracy.

\paragraph{Named Entity Recognition}
NER evaluation used the German CoNLL 2003 \citep{tjong_kim_sang_introduction_2003} and GermEval 2014 \citep{benikova_germeval_2014} datasets. CoNLL 2003 includes four entity types, while GermEval 2014 provides fine-grained categories and supports nested annotations. Both were evaluated using the $F_{1}$ score, with GermEval using an adapted metric accounting for label and span equality.

\paragraph{Text Classification}
We evaluated classification on GermEval 2018 \citep{risch_fine-grained_2018} (German tweet sentiment analysis) and 10kGNAD \citep{schabus_one_2017} (German news categorization). GermEval 2018 followed the data splits defined by \citet{chan-etal-2020-germans}, while 10kGNAD used a predefined 90\%-10\% train–test split, with 10\% of the training set further held out for validation. Both tasks were evaluated using the mean $F_1$ score.

\subsection{Model Properties}
Table~\ref{tab:params} lists the vocabulary sizes and total parameter counts of all models included in our evaluation. While most German BERT-style base models, such as GBERT\textsubscript{base}, dbmdzBERT, and GELECTRA\textsubscript{base}, contain approximately 110 million parameters, GeistBERT and \GottBERTf{base}{x} are slightly larger at around 126 million parameters due to their RoBERTa-based architecture and a larger vocabulary of 52,009 tokens.

Large-scale German models such as GBERT\textsubscript{large}, GELECTRA\textsubscript{large}, and \GottBERTf{large}{x} contain between 335 and 357 million parameters. Among the multilingual models, XLM-RoBERTa\textsubscript{base} and XLM-RoBERTa\textsubscript{large} are substantially larger, with 278 million and 560 million parameters respectively. The vocabulary sizes vary across models and are influenced by tokenizer design and pre-training data. GeistBERT uses the same tokenizer as GottBERT, which is based on byte-level BPE trained on German text.

\begin{table}[!htbp]
\caption{The size of the vocabulary and the size of the parameters are shown for the model types used in this study. This table does not show other design differences of the models. Values were extracted using Huggingface's transformers library.}\label{tab:params}
\centering\small
\begin{tabular}{lcc}%
    \hline
    \bfseries Model & \bfseries Vocab Size & \bfseries \#Params
    \\\hline
    \csvreader[late after line = \\]{params.csv}{}
     {\csvcoli\ & \csvcolii & \csvcoliii}
     \hline
\end{tabular}
\end{table}

\section{Results}
\subsection{Training Dynamics}
\begin{figure}[htb]
    \centering
    \includegraphics[width=\columnwidth]{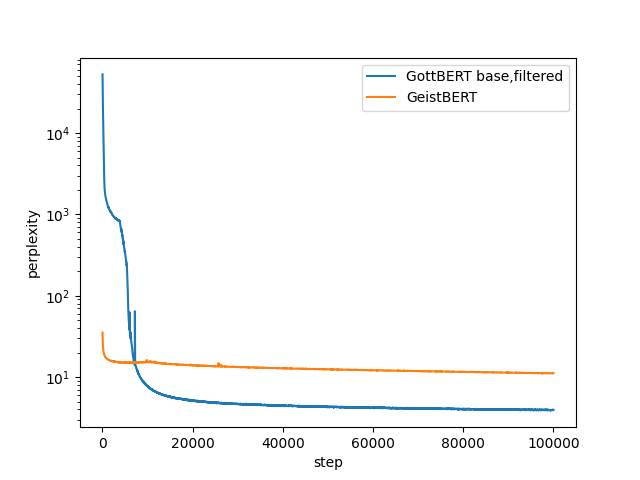}
    \caption{\label{fig:perplexity}Perplexity of {\GottBERTf{base}{x}} and GeistBERT, evaluated on the validation set after each optimization cycle; values are plotted on a logarithmic y-axis.}
\end{figure}

During the model pre-training the perplexity of the model is computed based on a test set for each optimization cycle (see Figure \ref{fig:perplexity}). After an initial sharp decrease, perplexity briefly increased for several steps before gradually declining until the final step. We assume that, given more training time, it would have continued to decrease further. The entire pre-training process required approximately 8.3 days of computation time.

Importantly, GeistBERT started from a relatively low perplexity due to continued pre-training. In comparison, {\GottBERTf{base}{x}} (trained entirely from scratch) started with a perplexity of about 52{,}592 and converged to around 4, whereas GeistBERT began at 35.17 and converged down to approximately 11. This illustrates the potential stability and efficiency benefits of continued pre-training in reaching useful representations quickly.

\subsection{Downstream Tasks}

GeistBERT sets a new state-of-the-art among base models for German NLP, outperforming all comparable models and closely approaching large-scale model performance across tasks. It even achieves absolute SOTA in GermEval 2018 fine-grained classification (see Table~\ref{tab:all}).

The optimal hyperparameters selected per task are summarized in Table~\ref{tab:best_params}, extending the original {\GottBERTt} setup \citep{scheible-etal-2024-gottbert} by including GeistBERT models. The total computation time for all downstream evaluations amounted to 517 hours and 24 minutes ($\approx$21.6 days) on two Nvidia RTX~3090 GPUs, detailed per task in Table~\ref{tab:gpu_time}.

\begin{table}[htb]
    \caption{Computation time in hours and minutes for the downstream tasks summing up to 517 hours and 24 minutes, which are approximately 21.6 days.}
    \label{tab:gpu_time}
    \centering
    \begin{tabular}{llc}
         \hline
         \bfseries Task & & \bfseries Computation Time\\
         \hline
         \multicolumn{2}{l}{XNLI} & 47:52 \\
         \multicolumn{2}{l}{GermEval 2014} & 235:36 \\
         \multicolumn{2}{l}{CoNLL03} & 92:45 \\
         \multirow{2}{*}{GermEval 2018} & coarse & 45:46 \\
          & fine & 43:25 \\
         \multicolumn{2}{l}{10kGNAD} & 85:49 \\
         \hline
    \end{tabular}
\end{table}

\begin{table*}[!htbp]
\caption{Hyperparameters of the best downstream task models for each task and pre-trained model. This table extends the original {\GottBERTt} setup by including GeistBERT models. BS refers to batch size, and LR denotes the learning rate.}
\label{tab:best_params}
\centering
\begin{tabular}{lcccccccccc}%
    \hline
    \multirow{2}{*}{\bfseries Model} & 
    \multicolumn{2}{c}{\multirow{2}{*}{\bfseries GermEval 2014}} & 
    \multicolumn{2}{c}{\multirow{2}{*}{\bfseries CoNLL 03}} & 
    \multicolumn{4}{c}{\bfseries GermEval 2018} & 
    \multicolumn{2}{c}{\multirow{2}{*}{\bfseries 10kGNAD}} \\
    & & & & & \multicolumn{2}{c}{\bfseries coarse} & \multicolumn{2}{c}{\bfseries fine} & &  
    \\\hline
      & BS & LR & BS & LR & BS & LR & BS & LR & BS & LR 
    \\\cmidrule{2-11}
    \csvreader[late after line = \\]{hyperparams_base.csv}{}
     {\csvcoli & \csvcolii & \csvcoliii & \csvcoliv & \csvcolv & \csvcolvi & \csvcolvii & \csvcolviii & \csvcolix & \csvcolx & \csvcolxi}
     \hline
    \csvreader[late after line = \\]{hyperparams_large.csv}{}
     {\csvcoli & \csvcolii & \csvcoliii & \csvcoliv & \csvcolv & \csvcolvi & \csvcolvii & \csvcolviii & \csvcolix & \csvcolx & \csvcolxi}
     \hline
\end{tabular}
\end{table*}

\paragraph{NLI}  
GeistBERT\textsubscript{base} achieves an accuracy of 82.67\% on the German NLI task, outperforming all other base models in our evaluation. While it does not surpass top-scoring large-scale models such as GELECTRA\textsubscript{large} (86.33\%) or GBERT\textsubscript{large} (84.21\%), it performs competitively and even surpasses \GottBERT{large}{x} (82.46\%) and nearly matches \GottBERTf{large}{last} (82.79\%), narrowing the performance gap despite its smaller size.

\begin{table*}[!htbp]
\caption{All the results of the experiments are shown in percent. They are all based on the test set and the best score out of 24 runs (selection based on validation set). While NLI is measured by accuracy, all the other metrics are $F_1$ measures. Per model size, best results are \textbf{bold}, second-best \underline{underlined}. Results for {\GottBERTt} are reported on both the unfiltered and filtered corpora, the latter indicated by $^\mathrm{f}$. For each {\GottBERTt} model, we include both the best and last checkpoint of the pre-training, with the last denoted by $^\crosssymbol$. Values for non-GeistBERT models are taken from \citet{scheible-etal-2024-gottbert}.}\label{tab:all}
\centering\small
\begin{tabular}{lccccccc}%
    \hline
    \multirow{2}{*}{\bfseries Model} & \multirow{2}{*}{\bfseries XNLI} & \multirow{2}{*}{\bfseries GermEval 2014} & \multirow{2}{*}{\bfseries CoNLL 03} & \multicolumn{2}{c}{\bfseries GermEval 2018} & \multirow{2}{*}{\bfseries 10kGNAD} \\
    &  &  & & \bfseries coarse & \bfseries fine &  
    \\\hline
    \csvreader[late after line = \\]{all_base.csv}{}
     {\csvcoli\ & \csvcolii & \csvcolv & \csvcolviii & \csvcolxi & \csvcolxiv & \csvcolxvii}
     \hline
    \csvreader[late after line = \\]{all_large.csv}{}
     {\csvcoli\ & \csvcolii & \csvcolv & \csvcolviii & \csvcolxi & \csvcolxiv & \csvcolxvii}
     \hline
     
\end{tabular}
\end{table*}

\paragraph{Named Entity Recognition}  
GeistBERT achieves strong $F_1$ scores on both CoNLL 2003 (86.17\%) and GermEval 2014 (88.47\%), outperforming all other base models in our evaluation. It also surpasses all large-scale {\GottBERTt} variants on GermEval 2014 and comes remarkably close on CoNLL 2003, with only a 0.11\% gap to the lowest-scoring large variant. While top-performing large models such as GBERT\textsubscript{large} (87.19\% on CoNLL) and XLM-R\textsubscript{large} (88.83\% on GermEval) remain ahead, GeistBERT narrows the performance gap significantly, demonstrating robust entity representation capabilities despite its compact size.

\paragraph{Text Classification}  
GeistBERT\textsubscript{base} achieves strong performance across all classification tasks, ranking first in GermEval 2018 fine-grained classification (66.42\%), second in 10kGNAD (90.89\%), and third in GermEval 2018 coarse (79.67\%). It consistently outperforms all other base models and surpasses several large-scale models, particularly in the fine-grained setting. The results indicate that GeistBERT performs competitively across diverse classification benchmarks, despite being a base-sized model.


\section{Discussion}

\subsection{Principal Findings}
The continued pre-training of GottBERT on a broader and partially deduplicated and filtered German corpus consisting of OSCAR23, OPUS, mC4, Wikipedia, and OpenLegal, together with the use of WWM, leads to clear improvements across multiple language modeling tasks. GeistBERT establishes a new state of the art among base models and achieves competitive results with larger models across multiple German NLP benchmarks.

\subsection{Training Considerations and Data Quality}
In contrast to the TPU-based training used for GottBERT, GPU training also enabled more flexible preprocessing, such as sentence-aware segmentation. This made it possible to preserve natural sentence structure during training, even when using fixed-length sequences. Nevertheless, hyperparameter tuning remains a crucial factor for achieving strong downstream performance \citep{dodge_fine-tuning_2020}. WWM contributed to improved tokenization, aligning with previous findings \citep{martin_camembert_2020, chan-etal-2020-germans}.
However, we did not perform a dedicated ablation study comparing WWM with standard subword masking, as this would have required training an additional baseline model. Nevertheless, the consistently strong downstream results of GeistBERT suggest that WWM contributed positively, in line with earlier findings.
Moreover, we were able to adopt a higher peak learning rate (0.0007), which may also have been facilitated by initializing from the \GottBERTf{base}{best} checkpoint.

While deduplication and filtering were applied to CulturaX, other subcorpora (e.g., OPUS, Wikipedia, OpenLegal) were only partially processed or left unfiltered. This means that some redundant or lower-quality data may still be present. Prior work suggests that models benefit from increased corpus diversity \citep{martin_camembert_2020}, and GeistBERT’s use of many different corpora likely contributed to its robustness. Additionally, vocabulary size plays a role in performance \citep{10.1145/3578707}, though ours remains well-optimized.

We did not perform ablation experiments per subcorpus, as this would have required multiple additional large-scale pre-training runs. Nevertheless, we expect that improvements are not only attributable to the sheer size of the training data (1.3 TB), but also to the increased heterogeneity of the sources. The OSCAR23+mC4 portion clearly contributed the majority of the volume, while smaller corpora such as OpenLegal, Wikipedia, and OPUS are likely to have increased linguistic and domain diversity. Prior findings from CamemBERT \citep{martin_camembert_2020} indicate that variance of a corpus matter and impacts downstream robustness, which suggests that the mix of sources in GeistBERT was similarly beneficial.

\subsection{Continued Pre-training and Outlook}
We chose to continue pre-training from {\GottBERTt} rather than training GeistBERT from scratch, as it is common practice with domain-specific adaptations \citep{10.1093/jamiaopen/ooac087, 10.1093/bioinformatics/btz682, digital2040030, DBLP:journals/corr/abs-2004-10964}. This allowed us to reuse German-specific tokenization and pre-trained weights, and to focus on training and evaluating a single, well-defined setup within time constraints. While training from scratch with a custom vocabulary may yield more tailored embeddings \citep{el-boukkouri-etal-2022-train}, prior work suggests that continued pre-training often achieves comparable results. A direct comparison between continued pre-training and training from scratch on the same architecture and corpus remains an interesting avenue for future work.

Following the broad adoption of GottBERT in German NLP \citep{scherrmann2023german, BRESSEM2024121598, 10.1093/jamiaopen/ooac087, xu2021bert, frei2022gernermed, FREI2023104478}, we hope GeistBERT will be similarly received and applied. 

\section{Conclusion}
In this work, we introduced GeistBERT, a German RoBERTa-based language model trained on a diverse as well as partially deduplicated and filtered corpus, incorporating WWM to enhance pre-training. GeistBERT achieves SOTA performance among base models and even outperforms several larger models across multiple tasks. These results underscore the importance of corpus diversity and WWM in improving downstream performance. GeistBERT is released under the MIT license on Huggingface, with fairseq checkpoints provided.

\section*{Limitations}
Several limitations should be acknowledged in this study. First, while deduplication and filtering were applied to CulturaX (OSCAR23 + mC4) and deduplication to selected OPUS corpora, other parts of the dataset (e.g., Wikipedia, OpenLegal) were not processed, potentially leaving redundant or noisy data.

Second, GeistBERT’s training data, though diverse, remains specific to the selected corpora (OSCAR23, OPUS, mC4, Wikipedia, OpenLegal). Its generalization to other datasets or domains remains uncertain, and performance on dialects and cultural nuances within German may be limited. Further fine-tuning could improve adaptability to regional language variations.

Third, we did not include a detailed error analysis of model predictions. While such an analysis could provide additional insights into systematic failure modes, our focus in this work was on efficiency and establishing strong baselines for German NLP.

Finally, due to efficiency constraints and limited computational resources, we did not train a large version of GeistBERT, as pretraining based on GottBERT estimates would have required approximately 4.75 times more compute. While our results demonstrate the strong performance of the base model, larger architectures could potentially achieve even better results.

\section*{Ethical Considerations}
Like all large-scale language models, GeistBERT may inherit biases from its training data, which can influence downstream tasks such as classification or decision-making. While deduplication reduces redundancy and noise, it does not remove deeper societal or representational biases. Furthermore, training on large web-based corpora raises privacy concerns, as models may inadvertently retain sensitive information. Responsible deployment is especially important in high-stakes domains like legal, medical, or financial NLP.

Despite optimizations for efficiency, pre-training and evaluating transformer models remain computationally demanding, contributing to energy use and carbon emissions. These environmental costs highlight the need for balancing model performance with sustainable development goals.


\bibliographystyle{acl_natbib}
\bibliography{anthology,ranlp2025}
\end{document}